\documentclass[letterpaper]{article}
\usepackage{style/aaai}
\usepackage{times}
\usepackage{helvet}
\usepackage{courier}
\usepackage{amsmath}
\usepackage[dvipsnames]{xcolor}
\usepackage{listings}
\usepackage[pdftex]{graphicx}
\usepackage{multirow}
\usepackage{amssymb}
\usepackage{float}

\title{Evaluating the quality of tourist agendas customized to different travel styles}

\author{Jes\'us Ib\'a\~nez-Ruiz, Laura Sebasti\'a and Eva Onaindia \\
Universitat Polit{\`{e}}cnica de Val{\`{e}}ncia\\
Camino de Vera s/n\\ E46022-Valencia (Spain)\\
jeibrui@dsic.upv.es, lstarin@dsic.upv.es, onaindia@dsic.upv.es
}

\begin{document}

\maketitle

\begin{abstract}
\begin{quote}

Many tourist applications provide a personalized tourist
agenda with the list of recommended activities to the user. These applications must undoubtedly deal
with the constraints and preferences that define the user interests. Among these preferences, we can find those that define the travel style of the user, such as the rhythm of the trip, the number of visits to include in the tour or the priority to visits of special interest for the user. In this paper, we deal with the task of creating
a customized tourist agenda as a planning and scheduling application capable of conveniently scheduling
the most appropriate goals (visits) so as to maximize the
user satisfaction with the tourist route.
This paper makes an analysis of the meaning of the travel style preferences and compares the quality of the solutions obtained by two different solvers, a PDDL-based planner and a Constraint Satisfaction Problem solver. We also define several quality metrics and perform extensive experiments in order to evaluate the results obtained with both solvers.

\end{quote}
\end{abstract}

\section{Introduction}

The exponential growth of the Internet of Things and the surge of open data platforms provided by city governments worldwide is providing a new foundation for travel-related mobile products and services. With technology being embedded on all organizations and entities, and the application of the smartness concept to address travellers' needs before, during and after their trip, destinations could increase their competitiveness level \cite{buhalis13}.

Many tourism applications provide a personalized tourist agenda with the list of recommended activities to the user \cite{Sebastia:2009,vansteenwegen11,RefanidisE15}. In most cases, applications return a tourist route or agenda indicating the most convenient order for the user to realize the activities and, additionally, the path to follow between activities. In other cases, tools provide a dynamic interaction that allows the user to interact with such agenda by adding or removing activities or changing their order.

{\sf SAMAP} \cite{castillo08}, for instance, elicits a tourist plan including information about the transportation mode, restaurants and bars or leisure attractions such as cinemas or theaters, all this accompanied with a detailed plan explanation. Scheduled routes presented in a map along with a timetable are nowadays a common functionality of many tourist applications like {\sf e-Tourism} \cite{Sebastia:2009}, including also context information such as the opening and closing hours of the places to visit and the geographical distances between places. In {\sf CT-Planner} \cite{kurata14}, personalization is understood as taking into account preferences like the walking speed or reluctance to walk of the user, in which case the planner will suggest short walking distances. {\sf PersTour}~\cite{LimCLK15} calculates a personalized duration of a visit using the popularity of the point of interest and the user preferences. The work in~\cite{RodriguezMPC12} considers user preferences based on the number of days of the trip and the pace of the tour; i.e., whether the user wants to perform many activities in one day or travel at a more relaxed pace.

Personalized tourism applications must undoubtedly deal with the constraints and preferences that define the interests of the user. This can be addressed through a scheduler designed for the automatic scheduling of user's individual activities in an electronic calendar such as {\sf SelfPlanner} \cite{RefanidisA11}. In {\sf SelfPlanner}, user preferences are defined over alternative schedules and the user specifies whether she prefers the task to be scheduled as early or as late as possible or whether she is indifferent on how a task will be scheduled within its temporal domain. {\sf SelfPlanner} and its descendent \cite{AlexiadisR16} enable users express their preferences over the way their activities should be scheduled in time. {\sf e-Tourism} \cite{Sebastia:2009,IbanezSO16} approaches the problem of creating a customized tourist plan as a preference-based planning problem encoded in PDDL3.0 and solved with the planner {\sf OPTIC} \cite{BentonCC12}. One limitation of {\sf OPTIC} is that it does not enable the use of non-linear plan metrics. This makes it unaffordable to deal with situations in which, for instance, the satisfaction of the user with tourist plan does not always increase linearly with its duration.

Tourist preferences are not only about scheduling activities at the user's preferred time but also dealing with the travel style or personal circumstances of the tourist such as the rhythm of the trip, handling the number of visits to include in the tour or giving more priority to visits of special predilection for the user. Thus, we envision the task of creating a customized tourist agenda as a planning and scheduling (P\&S) application capable of conveniently scheduling the most appropriate goals (visits) so as to maximize the user satisfaction with the tourist route. Our proposal relies upon
encoding the tourist agenda problem as a CSP (Constraint Satisfaction Problem). The challenge when using a CSP-based approach is specifically on (a) the encoding of a planning problem as a constraint programming formulation \cite{Garrido09,Sebastia:2009} and (b) the encoding of user preferences that must be maximized as a function to minimize.

The paper is organized as follows. First, a description of the tourist agenda problem is given and we detail the metrics we will use to evaluate each obtained solution. Then, we describe the formulation of this problem for being solved by an automated planner. Afterwards, the problem is encoded as a CSP. In the section {\em Experiments}, we analyze the results we have obtained for a set of tourist problems. Finally, we draw some conclusions about this work.

\section{Problem description}

In this section, we describe the tourist problem to be solved with a planner and a CSP solver. The problem is inspired in the tourist setting introduced in \cite{IbanezSO16}, which describes the problem of generating an agenda for a tourist. The information included in the problem definition is: (a) a set of recommended points of interest (POIs) for the particular tourist; we assume the existence of a Recommender System (RS) that returns a set of preferable POIs for the user according to her likes; (b) the tourist's preferences regarding the travel style and model of transport; and (c) context-ware information such as the location or hours of the tourist attractions.

\begin{figure*}
\centering
\includegraphics[width=18cm]{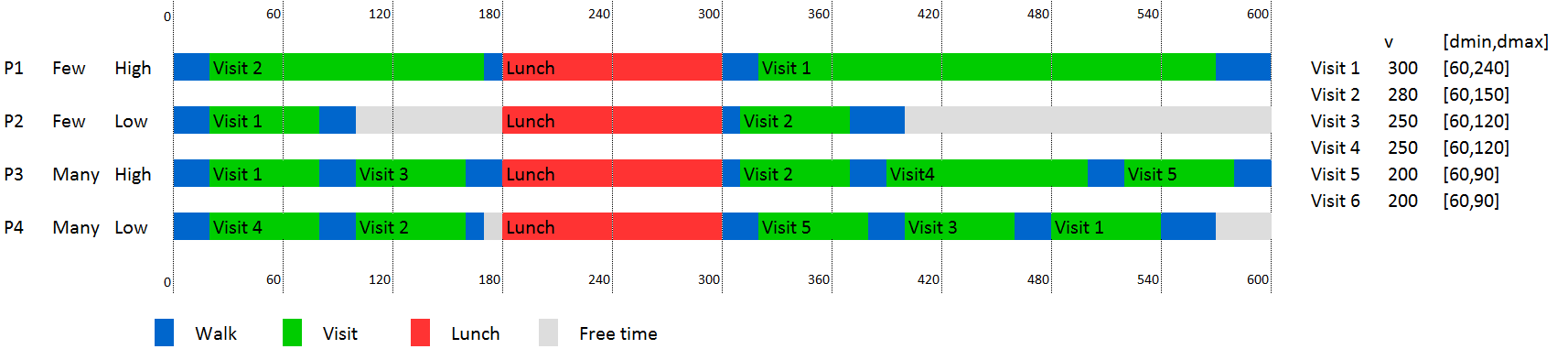}  
\caption{Example of agendas with different travel preferences ($vmax_p=300$)}
\label{plans}
\end{figure*}

Initially, the user enters the basic details of the route: the date of the visit, the start and finish point, start and finish hour, the time interval reserved for lunch and the mode of transport she prefers, which may determine the time needed to move between locations. Then, she also indicates her preferences related to her travel style, namely, she indicates if she prefers to include many or few visits in the tour or has
no preference over it and if she prefers to obtain an agenda with a high or a low temporal occupation or has no preference over it. Figure \ref{plans} shows four examples of agendas that reflect four combinations of these travel style preferences. The first one shows an agenda with a low number of visits but a high temporal occupation with no free time between visits. The second shows an agenda with a low occupation rate, where visits are shorter in order to have some free time between visits. The third example shows a high occupation rate agenda that also contains a high number of visits (in this case, visits 1 and 2 are shorter to be able to include visits 3, 4 and 5). The last example is an agenda with a low occupation rate but a high number of visits.

\subsection{Problem formulation}

In order to define a tourist problem, we need to distinguish between the domain data relative to the particular area or city the user wants to visit, and the user data, which specify the personalized features of the tourist route.

\begin{enumerate}
  \item Domain information. Two sources of knowledge from the domain are relevant for a tourist problem: the POIs of the city along with their opening and closing times and the travelling time between POIs.
      \begin{enumerate}
        \item Information about the POIs. For each POI $p$ of the city, we store three values: the POI identifier and the opening and closing time of $p$ (denoted by $open_p$ and $close_p$, respectively). Times are measured from 00:00. For example, $<$Cathedral, 600, 1170$>$ represents that the POI 'Cathedral' opens at 10:00; i.e., $open_{Cathedral}$=600 (10h*60min/h+0min = 600); and closes at 19:30; i.e., $close_{Cathedral}$=1170 (19h*60min/h+30min = 1170). The information of the hours of the city POIs will be denoted as $H^*$.
        \item Travelling times. Additionally, for every two POIs of the city, including the initial and final location of the user's route, we store the travelling time between them accordingly to the transport mode (e.g., walk, bus, car). For example, $<$Hotel, Cathedral, walk, 20$>$, $<$Hotel, Cathedral, bus, 8$>$. We will denote the data of travelling times between POIs by $T^*$.
      \end{enumerate}
  \item Personalized route information. We distinguish between the data that are directly provided by the user (route details) and the data estimated by the Recommendation System for the user:
      \begin{enumerate}
        \item Route details. The user introduces the following data:
        \begin{itemize}
          \item initial and final time of the route, which define the $total\_time$ available for the route
          \item initial and final location of the route, denoted by $start\_loc$ and $final\_loc$, respectively
          \item initial and final time for the lunch break, if specified
          \item the transport mode (e.g., walk, bus, car)
          \item preference of the user for the number of visits; the user can select among the values $\{few,many,indif\}$
          \item preference of the user for the temporal occupation of the route; the user can select among the values $\{high,low,indif\}$
        \end{itemize}

        We will refer to the route details introduced by the user as $R$.

        \item Recommendation. A set of recommendable POIs for the user to visit are obtainable through the RS. Specifically, for each recommended visit, the RS returns a tuple of the form $\langle p,v_p,dmin_p, dmax_p \rangle$, where $p$ is the POI to visit, $v_p \in [0,vmax_p]$  is the estimated value of $p$ to the user (i.e., the estimated degree of interest of the user in $p$), and $dmin_{p}$ and $dmax_{p}$ are the minimum and maximum recommended duration for visiting $p$, respectively. We will denote the set of recommended visits to a user as $V$\footnote{The RS we used to elicit the list of recommended POIs and values assumes $vmax_p=300$.}.
      \end{enumerate}
\end{enumerate}

A tourist problem $P^u$ for a particular user is defined as a tuple $P^u=<R,V,H,T>$,  where $R$ is the set of route details specified by the user, $V$ is the set of recommended visits to the user, $H \subseteq H^*$ is the hours of the POIs contained in $V$ and $T \subseteq T^*$ is the travelling times between the POIs in $V \cup \{start\_loc, final\_loc, restaurant\}$ according to the transport mode selected by the user.

A {\bf solution for a problem $P^u=<R,V,H,T>$} is a sequence of actions or plan $\Pi$ that contains \texttt{move} actions from $T$ and \texttt{visit} actions from $V$. More specifically, $\Pi=\{T_{\Pi},V_{\Pi}\}$, where:

\begin{itemize}
\item $T_{\Pi}$ are actions of the form $(\texttt{move} \; p \; q \; t_s \; dur_{p,q})$, being $p$ and $q$ two POIs, $t_s$ the start time of the move action and $dur_{p,q}$ the travelling time between $p$ and $q$ according to the user's selected transport.
\item $V_{\Pi}$ are actions of the form $(\texttt{visit} \; p \; t_s \; dur_p)$, being $p$ the POI to visit, $t_s$ the start time of the visit and $dur_p \in [dmin_p,dmax_p]$ the duration recommended for the visit. The $restaurant$ is also included in this set with $v_{restaurant}=0$\footnote{In this work, restaurants are not defined as POIs. We consider a generic POI \emph{restaurant} that the user must visit if a lunch break is specified. Including a list of restaurants as POIs of $V$ alongside their recommended value to the user is straightforward.}.
\end{itemize}

The goal is to maximize the user satisfaction; that is, including the most-valued recommended visits of $V$ and meeting the user preferences with respect to the number of visits and temporal occupation. We define a \textbf{penalty cost} for violation of user preferences and we pose the problem as finding the solution with minimal penalty.

\subsection{Penalties}

There can be many different solutions for the tourist problem defined above, given that different visits can be selected to be included in the plan, with different durations, in different order (which implies different movements from one location to another), etc. Each of these solutions will fit the user preferences in a better or worse way. The solver is aimed at finding the best plan according to some metrics that rely on four types of penalties to assess the degree to which the user preferences are not satisfied. Next, we present the four penalties, which are values in the interval [0,1].

\begin{itemize}

\item \textbf{Non-visited POIs:} This penalty is used to force the solver to include in the plan those visits with the highest recommendation value. That is, the idea is to obtain a plan with a high utility for the user, where {\em utility} is defined with respect to the recommendation value. We have designed three different utilities which, in turn, define three different penalties by substracting the obtained utility from the highest possible utility.
    Equation \ref{u1} calculates the utility as the ratio of the recommendation value of the visits included in the plan with respect to the maximum recommendation value that a plan could have, that is, the sum of the recommendation value of all the elements in $V$\footnote{Abusing the notation, we will write $p \in V$ or $p \in V_{\Pi}$ when we refer to the POI $p$ of an action {\tt visit} $p \; t_s \; dur_p$. And we will use $(p,q) \in T_{\Pi}$ when we refer to the POIs $p$ and $q$ of an action {\tt move} $p \; q \; t_s \; dur_{p,q}$).}:

\begin{equation}\label{u1}
U1 = \frac{\sum_{p \in V_\Pi} v_p}{\sum_{p \in V} v_p} \;\;\;\;\;\;\;\; P_{U1} = 1 - U1
\end{equation}

The second way to calculate this penalty is formalized in equation \ref{u2}. In this case, the utility is calculated as the sum of the recommendation value of all the visits included in the final plan multiplied by the time spent in each visit, with respect to the total time spent in the plan (that is, taking into account movements between locations). The aim of this penalty is to consider how long the user is visiting a place with a high recommendation value, not only that this place is visited.
\begin{equation}\label{u2}
U2=\frac{\sum\limits_{p \in V_\Pi}(v_p*dur_p)}{total\_time} \;\;\;\;\;
P_{U2} = \left(vmax_p - U2 \right) / vmax_p
\end{equation}

And finally, the third way to calculate this penalty, shown in equation \ref{u3}, is similar to equation \ref{u2}, but the recommendation value per time unit is divided by the total time spent on visiting POIs, instead of the total available time:
\begin{equation}\label{u3}
U3=\frac{\sum\limits_{p \in V_\Pi} (v_p*dur_p)}{\sum\limits_{p \in V_\Pi} dur_p} \;\;\;\;\;
P_{U3} = \left(vmax_p - U3 \right) / vmax_p
\end{equation}

For example, for the first plan in Figure \ref{plans}, where the recommendation value $v$ and the duration interval for each visit are shown on the right and $vmax_p=300$, the penalties would be calculated as follows:

\begin{itemize}
\item $P_{U1}=1-\frac{300+280}{1480}=0.61$, where 1480 is the sum of $v$ for all the visits.

\item $P_{U2}=(300-\frac{300*240+280*150}{600})/300=0.37$, where 240 and 150 are the duration of visits 1 and 2, respectively, and 600 is the total available time of the user.

\item $P_{U3}=(300-\frac{300*240+280*150}{240+150})/300=0.03$. This is a low value because, in fact, the recommendation value per time unit is nearly maximal.
\end{itemize}

If we compare these values of the penalties for the first plan, with the values obtained for the fourth plan, which are 0.14, 0.57 and 0.15, respectively, we can observe that: (1) $P_{U1}$ is much better for the fourth plan, because it includes three more visits; (2) however, $P_{U2}$ and $P_{U3}$ are better in the first plan, because the utility per time unit is higher as the new visits included in the fourth plan have a lower recommendation value.

\item \textbf{Movement time:} Movement time is the sum of all time needed to move between the locations included in the plan. The aim of this penalty is to force the solver to reduce the time spent in moving from one location to another.
\begin{equation*}\label{p_journey}
P_{journey}=\frac{\sum_{(p,q) \in T_\Pi} dur_{p,q}}{total\_time}
\end{equation*}

Using the first plan as in the example above, this penalty would be calculated as: $P_{journey}=\frac{20+10+30+20}{600}=0.13$


\item \textbf{Number of visits:} As for the preference regarding the number of visits, which denotes whether the user desires a tour with many POIs to visit, few POIs or it is indifferent to her, the penalty considers the number of visits included in the plan with respect to the total recommended places:
\begin{equation*}\label{p_visits}
P_{\#visits} =
\begin{cases}  
    \frac{|V| - |V_\Pi|}{|V|} & \text{if} \quad many\\
    0 & \text{if} \quad indif.\\
    \frac{|V_\Pi|}{|V|} & \text{if} \quad few\\
\end{cases}
\end{equation*}

Taking into account that the first plan in Figure \ref{plans} represents a plan with a few number of visits, this penalty would be calculated as: $P_{\#visits} = \frac{2}{6}=0.33$.
Obviously, the fourth plan with the same preference would have a higher value of the penalty, specifically 0.83, given that it includes 5 visits.

\item \textbf{Occupation:} If the user selects a high temporal occupation, the free time must be minimized. If the user selects a low temporal occupation, then the variables to minimize are the time spent on visits or travelling. In case the user selects ''indifferent'', it is not needed to minimize any expression. $free\_time$ is defined as the slack time between activities and it is calculated as the difference between the total available time and the time spent in actions in the plan:
\begin{equation*}
free\_time=total\_time - \sum\limits_{p \in V_\Pi} dur_p - \sum\limits_{(p,q) \in T_\Pi} dur_{p,q}
\end{equation*}

Therefore, the penalty is defined as follows:
\begin{equation*}\label{p_occup}
P_{occup} =
\begin{cases}
	\frac{free\_time}{total\_time} & \text{if} \quad high\\
     0 & \text{if} \quad indif.\\
     \frac{1}{free\_time*total\_time} & \text{if} \quad low\\
\end{cases}
\end{equation*}

Taking into account that the first plan in Figure \ref{plans} represents a plan with a high temporal occupation, this penalty would be calculated as: $P_{occup} = \frac{0}{600}=0$,
which is consistent with the fact that there is not free time in this case and, therefore, there is not any penalty due to occupation. In contrast, these penalty for the second plan would take a value of 0.47.

\end{itemize}

\subsection{Metrics}

In this section, we define the set of metrics that make use of the penalties introduced above. Given that these are penalties, a plan satisfies better the user preferences when the values of the penalties are lower; therefore, the metrics must be minimized.

In general, these metrics consist in the addition of the four penalties defined above. That is, all the factors are considered equally.

\begin{itemize}
\item \textbf{M1: utility per POI:} This metrics uses the first defined not-visited POIs penalty $P_{U1}$, therefore emphasis is put on how many POIs with high utility are visited. It will take a value in the interval [0,4].
\begin{equation*}\label{p_total}
M1=P_{U1}+P_{journey}+P_{\#visits}+P_{occup}
\end{equation*}

For example, assuming that the user prefers few visits and a high temporal occupation in the plan, this metrics would take the following values for each plan in Figure \ref{plans}: 1.09, 1.54, 1.18 and 1.28. This means that, according to this metrics, the best plan would be the first plan and the second plan would be the worst one; that is, in this particular example, the penalty due to the number of visits has not a great weight.

\item \textbf{M2: utility per time unit with respect to the total available time:} $M2$ uses the second not-visited POIs penalty $P_{U2}$, which considers the utility of each visit per time unit with respect the total time, thus taking into account the travelling actions, that do not provide any reward. Given that journeys are already considered, $P_{journey}$ penalty is excluded from $M2$, so this metrics will take a value in the interval [0,3]. Specifically:
\begin{equation*}\label{pm2}
M2=P_{U2}+P_{\#visits}+P_{occup}
\end{equation*}

For example, assuming the same preferences than above (few visits and a high temporal occupation), this metrics would take the following values for each plan in Figure \ref{plans}: 0.7, 1.61, 1.34 and 1.48; that is, the best plan would be the first plan and the second plan would be the worst according to this metrics. Again, the penalty for the number of visits has a lower impact than the penalty due to occupation in this example.

\item \textbf{M3: utility per time unit with respect to the time spent in visits:} 
 $M3$ considers the utility of the selected visits per time unit regarding the time spent in visits only, with the aim of focusing on how satisfying are the visits for the users, in spite of the time spent on travelling from one place to another. Specifically:
\begin{equation*}
M3=P_{U3}+P_{journey}+P_{\#visits}+P_{occup}
\end{equation*}

For example, assuming the same preferences than above (few visits and a high temporal occupation), this metrics would take the following values for each plan in Figure \ref{plans}: 0.51, 0.96, 1.2 and 1.28; that is, the best plan would be the first plan and the fourth plan would be the worst according to this metrics. Unlike the previous metrics, in this particular example, $P_{U3}$ is penalizing, in a greater degree, the plans with many visits than the plans with a low occupation.

\end{itemize}

In summary, we can observe that, in all cases, the metrics have selected correctly the best plan according to the user preferences. The degree in which the other plans are penalized depend on the particular configuration of the plan, as the results will show.

\section{PDDL encoding}

This section describes the planning formulation of the tourist problem described in the prior section. We will specify the problem with PDDL (Planning Domain Definition Language), the standard language for encoding planning problems.

The features required to define the tourist problem in PDDL are: (1) temporal planning and management of durative actions (e.g., duration of visits, time spent in transportation, etc.); (2) ability of reasoning with temporal constraints (e.g., scheduling the activities within the opening hours of places, planning the tour within the available time slot of the tourist, etc.) and (3) ability of reasoning with the tourist preferences (e.g., selecting the preferred activities of the user for planning the tour). Specifically, apart from durative actions, which were introduced in PDDL 2.1 \cite{fox2003pddl2}, we also need the following features:

\begin{itemize}
\item \textbf{Duration inequality} to define the duration of an action as a value within an interval. This functionality was included in PDDL2.1.
\item \textbf{Timed initial literals:} to describe deterministic and unconditional exogenous events. They were included in PDDL2.2 \cite{edelkamp04}.
\item \textbf{Preferences} or soft goals to express the user preferences. They were included in PDDL3.0 \cite{gerevini2009deterministic}.
\item \textbf{Plan metrics} to allow quantitative evaluation of plans for selecting the best plan. This was included in PDDL2.1.

\end{itemize}

Subsequently, we describe the PDDL problem formulation from the input data $P^u=<R,V,H,T>$.

\subsection{Variables}

The problem variables are specified through predicates and functions. Visiting a POI is described by means of:

\begin{itemize}
\item The interval duration of visiting a POI $p$ is defined through the functions {\tt (min\_visit\_duration ?p)} and {\tt (max\_visit\_duration ?p)}. They will be assigned the values $dmin_p$ and $dmax_p$ of the corresponding POI $p$ in the list $V$.
\item The opening and closing time of a POI $p$ are specified by a timed-initial literal: {\tt (at $open_p$ (open $p$))} and {\tt (at $close_p$ (not (open $p$)))}, where $open_p$ and $close_p$ are defined in $H$.
\end{itemize}

The time for moving from one location $p$ to another location $q$ is defined by the function {\tt (location\_time $p$ $q$)}, which indicates the time in minutes to move from $p$ to $q$ as indicated in $T$. We note that $T$ contains the travelling time between the POIs of $V$ according to the transport mode specified by the user. We also need the following predicates and functions:

\begin{itemize}
\item A predicate that represents the user initial location, {\tt (person\_at $start\_loc$)}, which will be modified when a \texttt{move} action is applied.
\item The function {\tt (free\_time)} represents the remaining available time; the initial value is set to $total\_time$, which will decrease as new activities are included in the plan.
\item Two functions to compute the metrics during the plan construction, namely, a function to count the number of visits included in the plan, {\tt (number\_visit\_location)}; and the function {\tt (transport\_time)} to add up the time spent in \texttt{move} actions.
\end{itemize}

\subsection{Actions}

\begin{figure}[t]
\begin{scriptsize}
\begin{verbatim}
(:durative-action move
  :parameters (?x - location ?y - person ?z - location)
  :duration (= ?duration (location_time ?x ?z))
  :condition
    (and
      (at start (person_at ?y ?x))
      (at start (>= (free_time)(location_time ?x ?z))))
  :effect
    (and
      (at start (not (person_at ?y ?x)))
      (at end (person_at ?y ?z))
      (at end (decrease (free_time)
        (location_time ?x ?z)))
      (at end (increase (transport_time)
        (location_time ?x ?z)))))
\end{verbatim}
\end{scriptsize}
\caption{Action {\tt move} of the tourism domain} \label{fig:move}
\end{figure}

\begin{figure}
\begin{scriptsize}
\begin{verbatim}
(:durative-action visit
  :parameters (?x - location ?y - person)
  :duration
    (and
      (>= ?duration (min_visit_time ?x))
      (<= ?duration (max_visit_time ?x))
      (<= ?duration (free_time)))
  :condition
    (and
      (at start (not_visit_location ?x))
      (over all (person_at ?y ?x))
      (over all (open ?x)))
  :effect
    (and
      (at start (not (not_visit_location ?x)))
      (at end (visit_location ?x))
      (at end (increase (number_visit_location) 1))
      (at end (decrease (free_time) ?duration))))
\end{verbatim}
\end{scriptsize}
\caption{Action {\tt visit} of the tourism domain} \label{fig:visit}
\end{figure}

We define three three types of actions in the tourist problem: \texttt{move}, \texttt{visit} and \texttt{eat} actions.

The action to \texttt{move} from one location to another is shown in Figure \ref{fig:move}. The parameters are the initial place {\tt ?x}, the user {\tt ?y} and the destination {\tt ?z}. The action duration is set to the time specified in $T$. The preconditions for this action to be applicable are: (1) the user is at location {\tt ?x} and (2) the free time is greater than the \texttt{move} duration. The effects of the action assert that (1) the user is not longer at the initial location, (2) the user is at the new location at the end of the action and (3) the free time and the time spent in the movement are modified accordingly to the action duration.

The action to \texttt{visit} a POI is defined in Figure \ref{fig:visit}. The parameters are the POI to visit {\tt ?x} and the user {\tt ?y}. The action duration is a value between {\tt (min\_visit\_time ?x)} and {\tt (max\_visit\_time ?x)} which must be smaller than the remaining available time {\tt (free\_time)}. The planner will choose the actual duration of the action according to these constraints. The conditions for this action to be applicable are: (1) the POI has not been visited yet; (2) the user is at the POI during the whole execution of the action; and (3) the place is open during the whole execution of the action. The effects of the action assert that (1) the POI has been visited\footnote{Two predicates are necessary to indicate a visit has been done if the planner does not allow for negated conditions.}, (2) the number of visited locations is increased and (3) the free time is updated according to the visit duration.

Finally, the \texttt{eat} action to represent the activity of {\em "having lunch"} is similarly defined to the \texttt{visit} action. 

\subsection{Goal and optimization function}

We define two different types of goals:

\begin{itemize}
\item {\em Hard goals} that represent the realization of an action that the user has specified as mandatory (e.g., the location the user has indicated final destination: {\tt (person\_at $location_{final}$)}).
\item {\em Soft goals or preferences} that we wish to satisfy in order to generate a good plan but that do not have to be achieved in order for the plan to be correct \cite{gerevini2009deterministic}. We will assign penalties to violated preferences.
\end{itemize}

The objective is to find a plan that achieves all the hard goals and minimize the total penalty for unsatisfied preferences. For example, the specification of metric $M1$ (see Sections \emph{Penalties} and \emph{Metrics}) in PDDL is expressed as:

\texttt{(:metric minimize (+ \indent \indent \indent $ P_{U1}  \; P_{journey} \; P_{\#visits} \; P_{occup}$))}

\vspace{0.1cm}

\textbf{(1)}$\;P_{U1}$. The specification of $P_{U1}$ requires defining a preference for every POI in $V$; e.g.

\vspace{0.1cm}

{\small
\texttt{(preference p1 (visit\_location id\_1))}

\texttt{(preference p2 (visit\_location id\_2))}}

\vspace{0.1cm}

and defining the penalty $P_{U1}$ for each preference:

{\small

\texttt{(/ (* 250 (is-violated p1)) 532)}

\texttt{(/ (* 282 (is-violated p2)) 532)}}

\vspace{0.1cm}

where $v_{id_1}=250$, $v_{id_2}=282$ and $\sum_{p \in V} v_p = 532$ (assuming $V$ contains only two POIs, {\small \texttt{id\_1}} and {\small \texttt{id\_2}}).

\vspace{0.1cm}
\textbf{(2)}$\,P_{journey}$. This penalty is specified as {\small \texttt{(/ (transport\_time) 540)}} where $total\_time=540$ in this particular example.

\vspace{0.1cm}
\textbf{(3)}$\;P_{\#visits}$. If the user selected \emph{few} visits, this penalty is expressed as {\small \texttt{(/ (number\_visit\_location) 10)}}, where $|V|=10$ in this example.

\vspace{0.1cm}
\textbf{(4)}$\;P_{occup}$. Assuming the user selected a \emph{high} temporal occupation in the plan, this is expressed as {\small \texttt{(/ (free\_time) 540)}} with $total\_time=540$.

\vspace{0.1cm}

\section{CSP model}\label{CSP}

This section details the specification of the tourist problem $P^u=<R,V,H,T>$ as a CSP.

\subsection{Constraints}

In this section we explain the constraints that it is necessary to specify in order to correctly solve the tourist problem.

\subsubsection{Plan structure.}

Among the $V$ places recommended by the RS, not all of them will be possibly included in the agenda due to several temporal restrictions. We define an array $P$ of $|V|+3$ elements that is used to record the places that will be included in the tourist route. The $|V|$ variables take a value in $\{0,1\}$ to denote whether or not the respective place is a visit to realize in the route. The three extra variables defined in $P$ represent the initial location of the user (always set to 1), the restaurant (this variable equals 1 if the user selected a lunch time interval) and the destination (always set to 1).

For example, for the first plan of Figure \ref{plans}, and assuming $|V|=6$, the final array $P$ will be $P = \left< Orig, Vis1, Vis2, Vis3, Vis4, Vis5, Vis6, Rest, Dest \right>$, where variables $P_0, P_1, P_2, P_7$ and $P_8$  are set to 1 and the value of variables from $P_3$ ($Vis3$) to $P_6$ ($Vis6$) is 0 because they are not included in the plan.

\subsubsection{Plan sorting.}

The constraints explained in this subsection are devoted to obtain a correct plan from the point of view of the ordering of the visits. Specifically, assuming that each visit included in a plan is assigned a number in the sequence, a plan is {\em correctly ordered} if: the current location of the user is assigned the 0th position and the destination of the user is assigned the nth position (if the plan has n+1 visits). This implies that there are not {\em empty positions} in the plan.

In order to obtain a correct plan, several additional structures and constraints must be added. Let $A_{ij}$ be a 2-dimension matrix (($|V|+3) \times (|V|+3)$) with components in a $\{0,1\}$ domain. $A_{ij}$ is used to represent the sequence of visits in the plan, where $i$ is the visited place and $j$ is the order of $i$ in the sequence.
For example, the first plan in Figure \ref{plans} would be stored in a matrix $A$ of $9x9$, where all the elements $A_{ij}$ are equal to 0, except $A_{00}$, which indicates that the user is initially at $start\_location$; $A_{21}$ to represent that the first to visit is $Vis2$; $A_{72}$ to indicate that next the user heads to the restaurant; $A_{13}$ to represent that in the next step the user visits $Vis1$ and $A_{84}$ to indicate that the user finishes her route in the destination.

$A_{ij}$ must fulfill two conditions: a place can only be visited once and two or more monuments cannot be visited at the same time.
\begin{equation*}
\forall i \sum_{j=0}^{|V|+2}A_{ij} \leq 1
\quad \quad \quad \quad \quad
\forall j \sum_{i=0}^{|V|+2}A_{ij} \leq 1
\end{equation*}

Let $m_{ijk}$ be a 3-dimension matrix ($(|V|+3) \times (|V|+3) \times (|V|+2)$) whose components take a value in $\{0,1\}$. This matrix establishes a relationship between a place to visit and the next one. That is, $m_{ijk}$ is set to 1 if place $i$ is visited immediately before $j$, and $i$ is visited in position $k$. More formally:
\begin{multline*}
\forall i,j,k / i,j \in [0, |V|+2], k \in [0, |V|+1] : \\ m_{ijk} = A_{ik} \ast  A_{j(k+1)}
\end{multline*}

Considering the matrix $A$ above, all the elements in the matrix $m_{ijk}$ are 0 except $m_{020}, m_{271}, m_{712}, m_{183}$. For example, $m_{271}$ denotes that after visiting the second place of array $P$ ($Vis2$) in position 1 the user heads to the fifth place of $P$ ($Rest$).

The aforementioned constraints may not prevent the model from generating an incorrect solution like the one shown before. In order not to have \emph{empty positions} in the sequence of visits, i.e., positions in which no visit is applied, we use the following constraint:
\begin{equation*}
\forall j \in [1, |V|+2]: \sum_{i=0}^{|V|+2}A_{ij} \leq \sum_{i=0}^{|V|+2}A_{i(j-1)}
\end{equation*}

The last place in the tourist agenda must be the user destination. Hence, if the user destination appears in the $j^{th}$ position of the sequence of visits, all the values  on the right of column $j^{th}$ in matrix $A$ must be 0.
\begin{equation*}
\text{If} \; A_{(|V|+2)j} = 1 \Rightarrow \forall z / j < z \leq |V|+2 : \sum_{z=0}^{|V|+2} A_{iz}=0
\end{equation*}

\subsubsection{Temporal constraints over the visits included in the plan}

These constraints are used to determine the value of $(start_i, finish_i)$ for each visit and lunch action $i$ included in the plan.

A recommended interval of the duration of each visit $[dmin_i, dmax_i]$ is provided by the RS. The following constraint establishes that the actual duration of a visit must fall within the recommended interval:
\begin{multline*}
\forall i \in [1,|V|+1] / P_i=1 : dmin_i \leq duration_i \leq dmax_i
\end{multline*}

The finish time of a visit is specified as:
\begin{equation*}
\forall i \in [1,|V|+1] / P_i=1 : finish_i = start_i + duration_i
\end{equation*}

A visit to a POI $i$ must be performed within the interval of the opening hours in $H$, denoted as $[open_i, close_i]$:
\begin{multline*}
\forall i \in [1,|V|+1] / P_i=1 : \\ open_i \leq start_i \wedge finish_i \leq close_i
\end{multline*}

In order to calculate the start time of a visit $j$, the estimated time needed to move from the prior visit $i$ (defined as $dur_{i,j}$ in $T$) must be taken into account:
\begin{multline*}
\forall i,j,k / i,j \in [0, |V|+2], k \in [0, |V|+1]: \\ \text{if} \; m_{ijk} = 1 \Rightarrow start_j > finish_i + dur_{i,j}
\end{multline*}

\subsection{Optimization function}

The constraints specified in the previous section enables the CSP solver to obtain a valid plan. Since our aim is to obtain a high-quality plan that fits the user's preferences, some other factors must be considered. In this case, we have implemented all the designed metrics as defined in Section {\em Metrics} by using the variables defined above. For example, the $P_{journey}$ penalty can be formalized as follows:

\begin{equation}
P_{journey} = \frac{\sum_{\forall i,j,k} dur_{i,j} * m_{ijk}}{total\_time}
\end{equation}

where $dur_{i,j}$ is the duration of the travelling time from $i$ to $j$ as defined in $T$.

\begin{figure*}[htb]
    \centering
    \begin{minipage}{.3\textwidth}
        \centering
        \includegraphics[width=5.5cm]{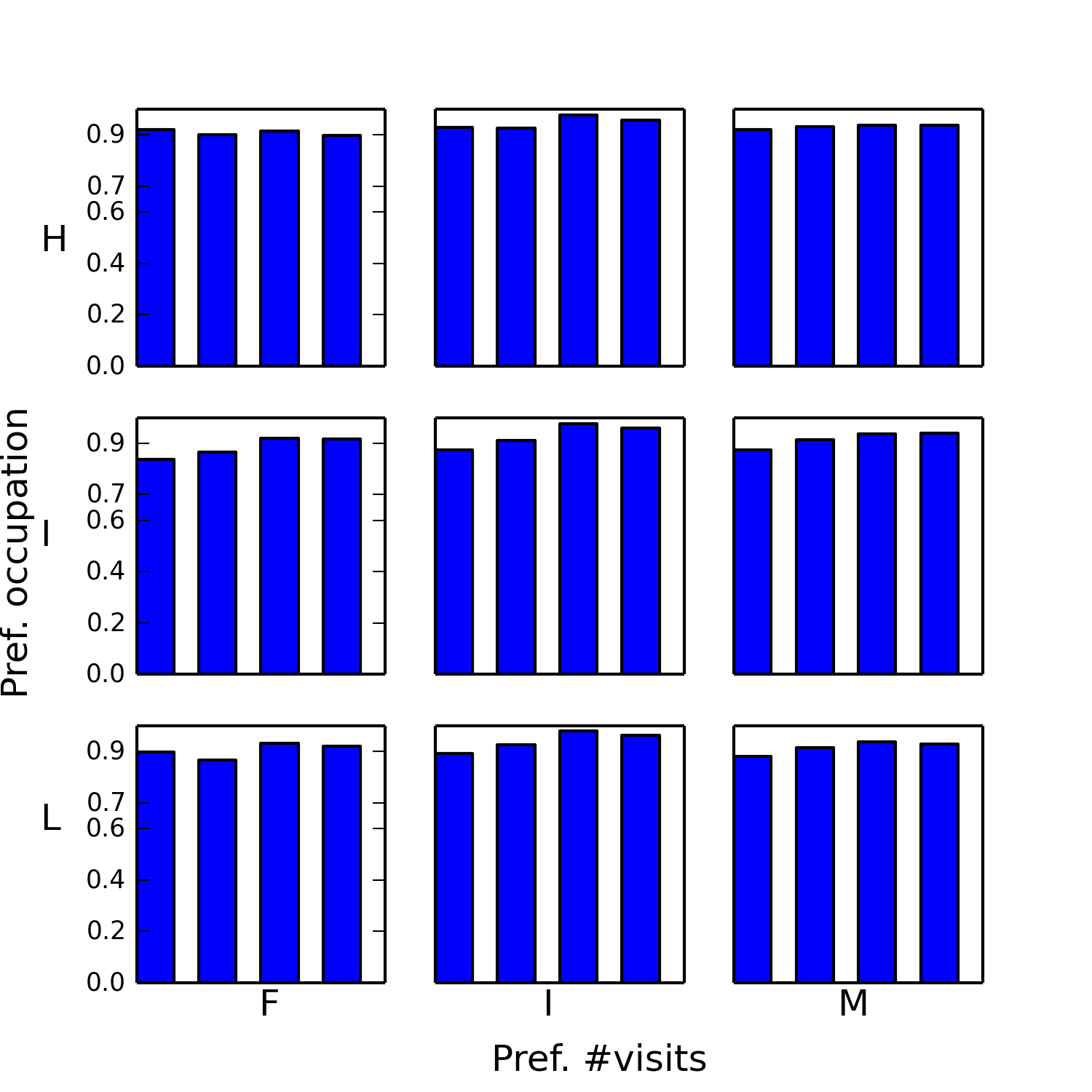}
        \caption{U1*}
        \label{fig:u3}
    \end{minipage}%
    \begin{minipage}{0.3\textwidth}
        \centering
        \includegraphics[width=5.5cm]{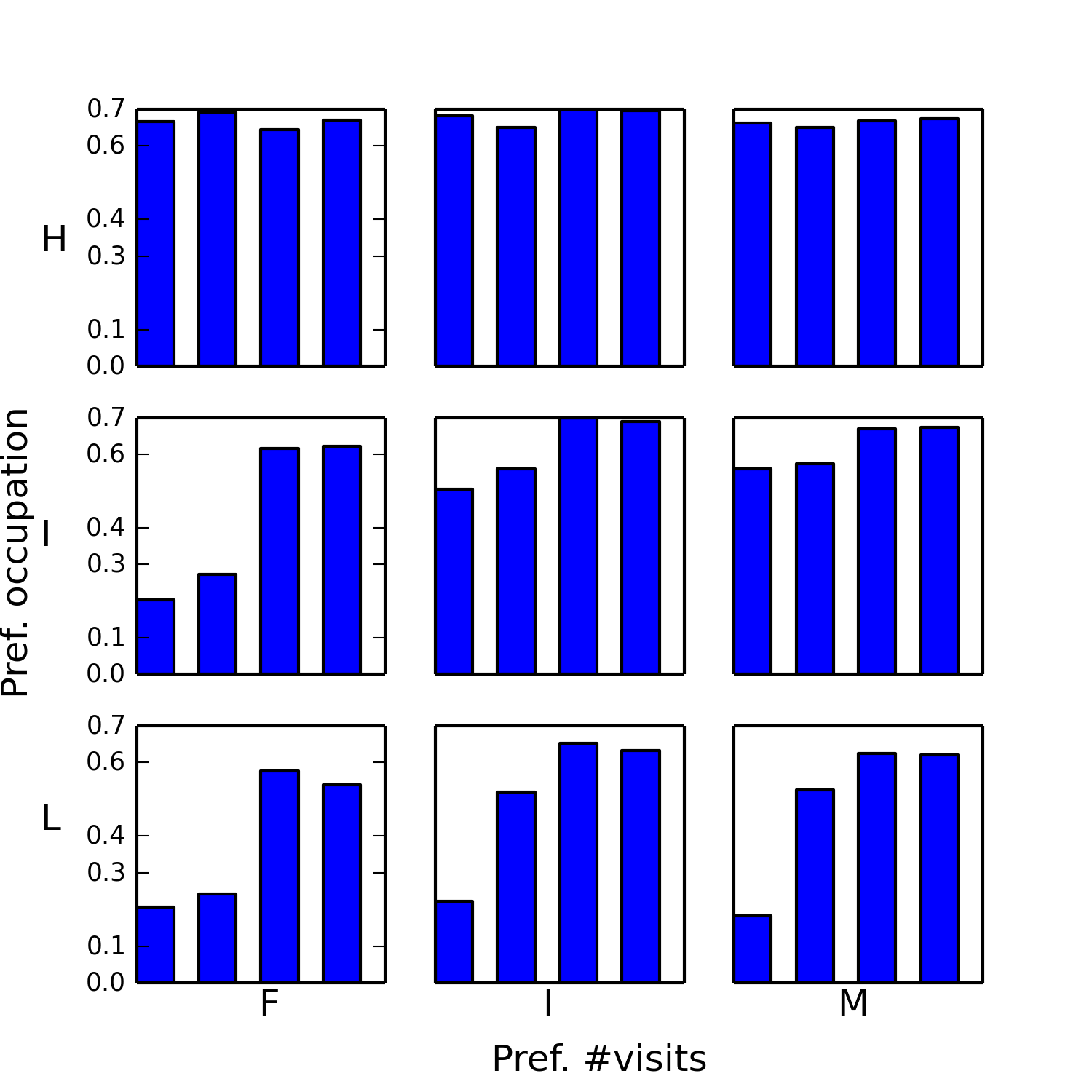}
        \caption{U2}
        \label{fig:u2}
    \end{minipage}
    \begin{minipage}{0.3\textwidth}
        \centering
        \includegraphics[width=5.5cm]{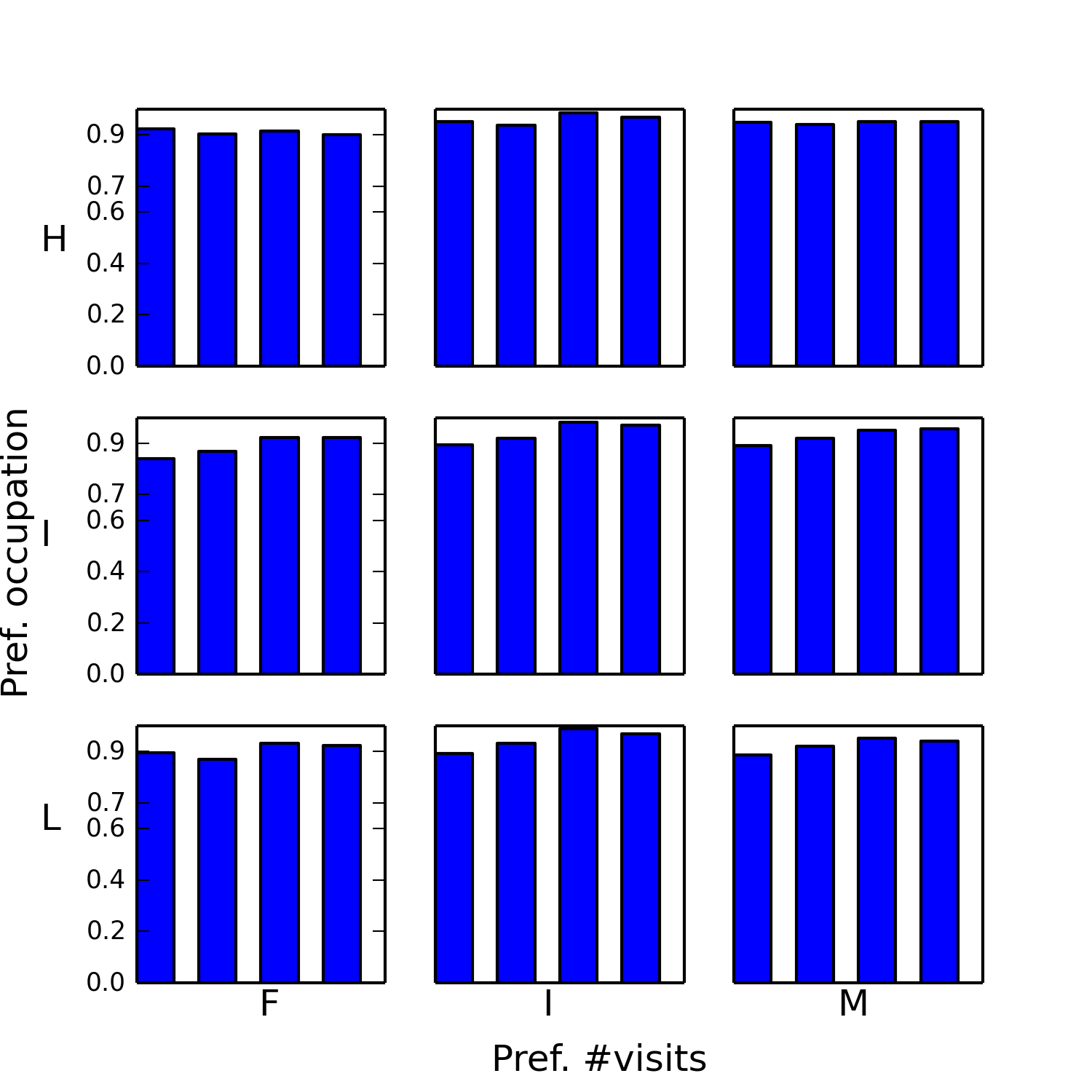}
        \caption{U3}
        \label{fig:u1}
    \end{minipage}
\end{figure*}

\begin{figure*}[htb]
\vspace{-0.4cm}
    \centering
    \begin{minipage}{.3\textwidth}
        \centering
        \includegraphics[width=5.5cm]{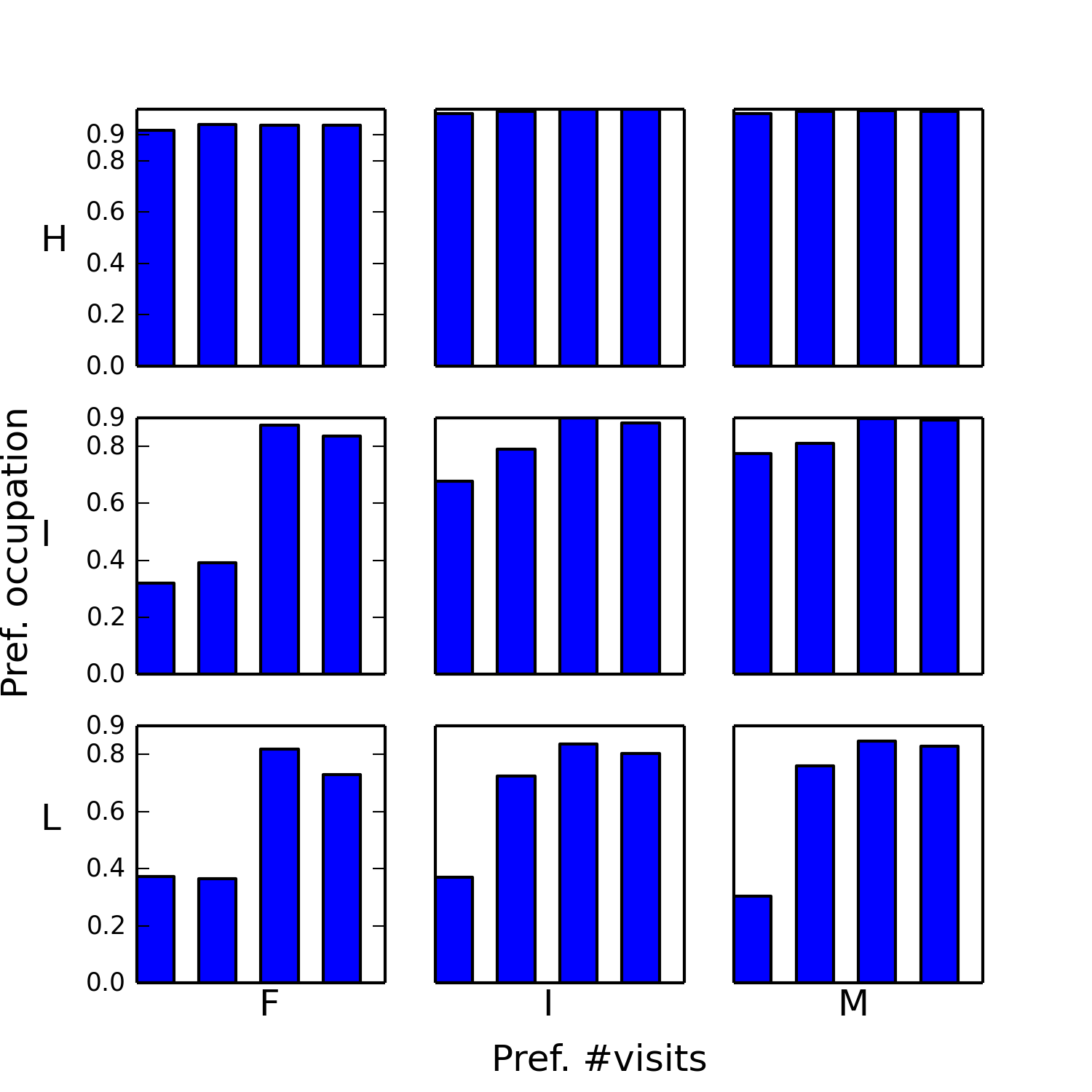}
        \caption{Percentage of occupation}
        \label{fig:occup}
    \end{minipage}%
    \begin{minipage}{0.3\textwidth}
        \centering
        \includegraphics[width=5.5cm]{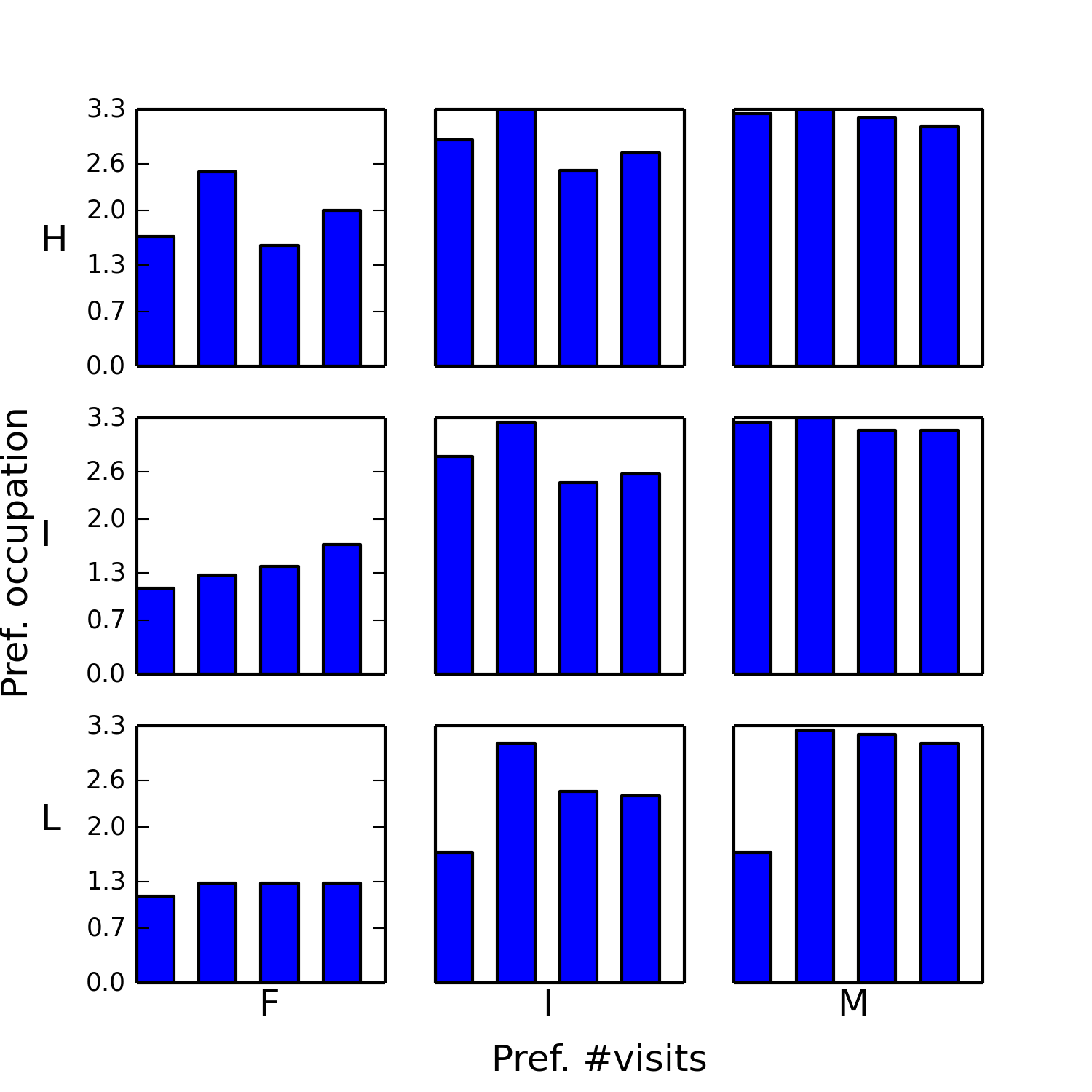}
        \caption{Number of visits}
        \label{fig:visits}
    \end{minipage}
    \begin{minipage}{0.3\textwidth}
        \centering
        \includegraphics[width=5.5cm]{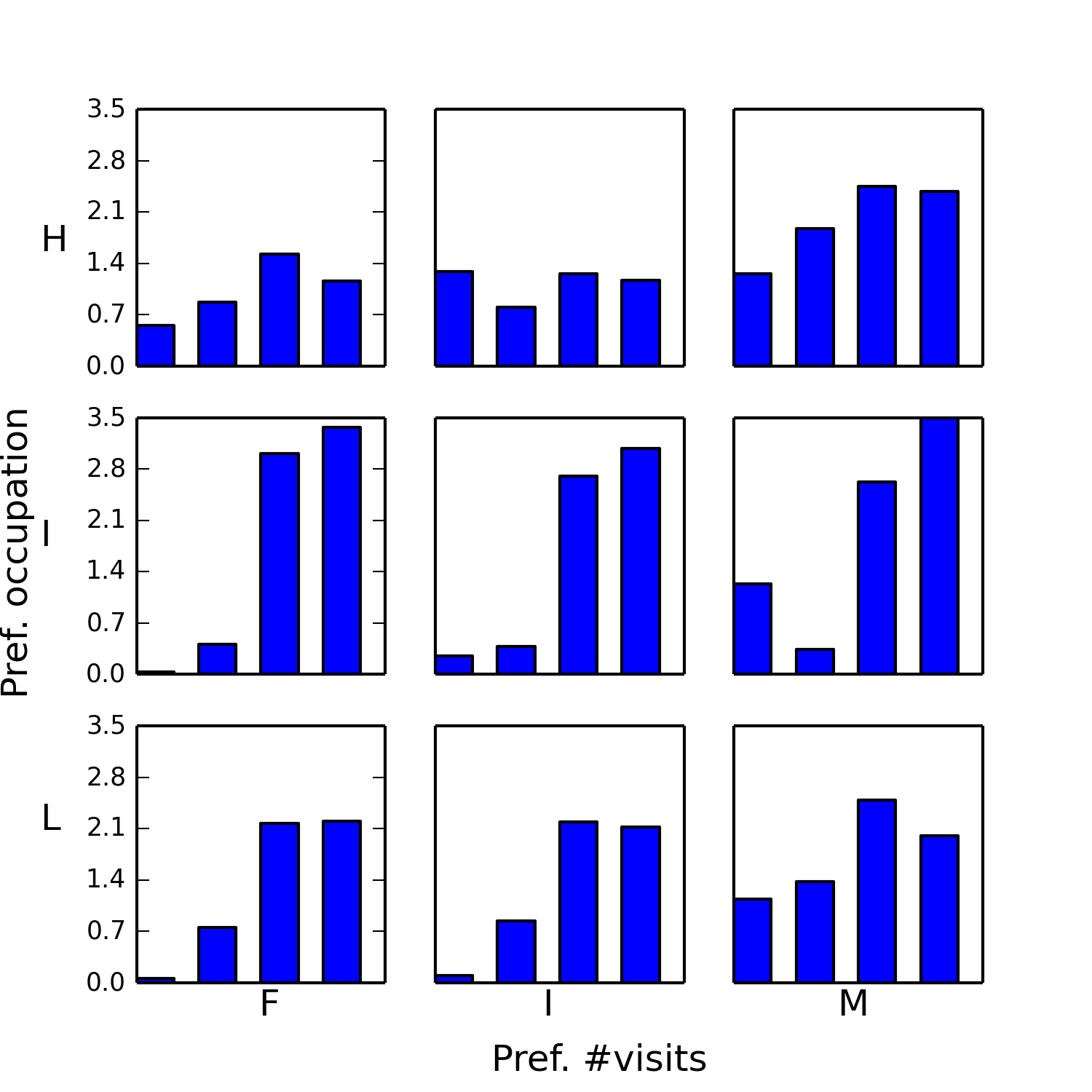}
        \caption{Execution time}
        \label{fig:execution}
    \end{minipage}
\end{figure*}

\subsection{Solution}

A CSP solution is a pair of the form $(start_i, finish_i)$ for the lunch action and for each recommended visit $i$. In case that the POI is included in the plan, $start_i$ and $finish_i$ indicate the start and finish time of the visit, otherwise these values are set to 0. Therefore, $V_{\Pi}$ would contain all the visits with values greater than 0 in $start_i$ and $finish_i$. On the other hand, each travelling action to be included $T_{\Pi}$ would be obtained from the gaps between visits.

The following table represents the CSP output from the first plan in Figure \ref{plans}. In this case, visits $V2$ and $V3$ and the restaurant would be included into $V_\Pi$ and, for example, the travelling action from the start location to the first visit ($V2$) would start at time 0 with a duration of 20.
\begin{center}
\begin{tabular}{l|l|l|l|l|l|l|l|}
\cline{2-8}
                                               & \textbf{V1} & \textbf{V2} & \textbf{V3} & \textbf{V4} & \textbf{V5} & \textbf{V6} & \textbf{R} \\ \hline
\multicolumn{1}{|l|}{\textbf{Start}} & 320         & 20          & 0           & 0           & 0           & 0           & 180        \\ \hline
\multicolumn{1}{|l|}{\textbf{Finish}}       & 570         & 170         & 0           & 0           & 0           & 0           & 300        \\ \hline
\end{tabular}
\end{center}

\section{Experiments}

This section presents the results obtained when solving the tourist problem with the PDDL-encoding and the CSP-encoding.

According to the features required to formulate the PDDL problem, we need a planner that handles PDDL3.0. A few automated planners are capable of this, such as {\small{\sf MIPS-XXL}} \cite{Gerevini06-mips}, {\small{\sf SGPLAN5}} \cite{chen05} or {\small{\sf OPTIC}} \cite{BentonCC12}. We opted for {\small{\sf OPTIC}} because it manages non-fixed durations, preferences and other helpful functionalities. {\small{\sf OPTIC}} is a heuristic planner that adapts the FF's relaxed-plan heuristic \cite{Hoffmann03} to temporal settings and numeric preference satisfaction. {\small{\sf OPTIC}} uses a hill-climbing algorithm in combination with a greedy algorithm, which enables to efficiently obtain good-quality solutions.

However, {\small{\sf OPTIC}} presents a severe limitation as it is not able to handle nonlinear functions. This prevents us from testing the metric $M3$ that involves the nonlinear penalty $U3$. On the other hand, let's assume the preference {\small\texttt{(preference p1 (visit location id\_1))}} in a problem; given that preference violation is expressed in PDDL through the variable {\small\texttt{(is-violated p1)}}, encoding $U2$ would require to be able to express {\small\texttt{(* (is-violated p1) (duration\_visit id\_1))}}, which is not allowed in {\small{\sf OPTIC}}. This nonlinearity restriction also affects the definition of the low temporal occupation penalty, reason why we implemented the occupation formula presented in \cite{IbanezSO16} where $P_{occup} = \frac{free\_time}{total\_time}$ if $high$ and $P_{occup} = \frac{total\_time - free\_time }{total\_time}$ if $low$.
In summary, {\small{\sf OPTIC}} has been tested in a modified version of metric $M1$ which we will refer to as $M1'$.

For solving the CSP-encoding, we chose a fast CSP solver like {\small{\sf CHOCO}} \cite{choco}. We used the {\small{\sf CHOCO}} function that, according with the manual, returns the optimal solution provided that no stop criteria is applied in the search. The variable and value selector is based on ``DowOverWDeg + LB'' strategy that solves the hard-constraints firstly and it reaches a good solution in a short time. Unlike {\small{\sf OPTIC}},  {\small{\sf CHOCO}} provides a greater flexibility and the possibility of testing all the defined metrics.

We tested the quality of the solutions obtained with the four possibilities ($M1'$ in {\small{\sf OPTIC}}, denoted as OPT-M1', and from $M1$ to $M3$ in {\small{\sf CHOCO}}, denoted by CHO-M1 to CHO-M3), as well as the temporal performance of both solvers\footnote{Experiments were performed in an intel i7-4790 3.6 Ghz machine with 16 GB DDR3.}. For doing this, we randomly generated a set of problems, with the following parameters: (1) a random $v_p$ value for each POI in the interval $[180,300]$; (2) a random distance between every two locations $T$ in the interval $[1,60]$; (3) for the visit duration, we take a random value between $30$ and $200$ minutes which determines the average duration visit (in case this value is larger than the total available time, the system will discard it); then, we compute the duration interval as explained in \cite{IbanezSO16} (i.e. we apply a normal distribution to obtain $dmin_p$ and $dmax_p$).

\begin{table}[tbp]

\centering
\caption{Average results wrt. the preference of occupation}
\label{tab:occup}
\begin{tabular}{|l|l|l|l|l|l|}
\hline
\multirow{2}{*}{\begin{tabular}[c]{@{}l@{}}Pref\\ Occup\end{tabular}} & \multirow{2}{*}{Metric} & \multicolumn{4}{l|}{Results}          \\ \cline{3-6}
                                                                      &                                & Occup & U1* & U2 & U3 \\ \hline
\multirow{4}{*}{High}                                                 & OPT/M1'                        & 0.924 & 0.881 & 0.669 & 0.897           \\ \cline{2-6}
                                                                      & CHO/M1                         & 0.937 & 0.876 & 0.662 & 0.885           \\ \cline{2-6}
                                                                      & CHO/M2                         & 0.940 & 0.899 & 0.669 & 0.907           \\ \cline{2-6}
                                                                      & CHO/M3                         & 0.939 & 0.888 & 0.678 & 0.896           \\ \hline
\multirow{4}{*}{Indif.}                                               & OPT/M1'                        & 0.614 & 0.822 & 0.421 & 0.834           \\ \cline{2-6}
                                                                      & CHO/M1                         & 0.690 & 0.855 & 0.468 & 0.860           \\ \cline{2-6}
                                                                      & CHO/M2                         & 0.919 & 0.900 & 0.660 & 0.908           \\ \cline{2-6}
                                                                      & CHO/M3                         & 0.898 & 0.895 & 0.660 & 0.905           \\ \hline
\multirow{4}{*}{Low}                                                  & OPT/M1'                        & 0.360 & 0.850 & 0.204 & 0.851           \\ \cline{2-6}
                                                                      & CHO/M1                         & 0.640 & 0.860 & 0.427 & 0.864           \\ \cline{2-6}
                                                                      & CHO/M2                          & 0.860 & 0.906 & 0.617 & 0.912           \\ \cline{2-6}
                                                                      & CHO/M3                          & 0.813 & 0.895  & 0.596 & 0.900          \\ \hline
\end{tabular}
\end{table}

We set three different values for the number of recommended places, $|V|=5$, $|V|=7$ and $|V|=10$, and three different values for the $total\_time$ of the route: a short plan (3 hours), half-day (5 hours) and all-day (9 hours). This makes a total of 9 different problem types. Only in the case of all-day routes, the plan will include the lunch time interval, which is a fixed-duration interval in a fixed time slot. Then, we added the 9 possible combinations of user preferences $P_{\#visits}=\{Many, Indifferent, Low\}$ and $P_{occupation}=\{High, Indifferent, Low\}$ to each problem type, thus having a total of 81 combinations. We generated two problem instances for each combination creating a total of 162 problems with a wide range of opening hours for each place.

We used different measures to evaluate the quality of the obtained plans with respect to the user preferences. We analyzed the number of visits in the plan and the level of occupation of the plan, that is, the ratio of the time the user is doing any activity (a visit, a journey between two places or having lunch) to the $total\_time$:

\begin{equation*}
Occup=1-\frac{free\_time}{total\_time}
\end{equation*}

Moreover, we analyzed the utility of the obtained plans with the utility measures $U1$, $U2$ and $U3$ defined in Equation \ref{u1}, \ref{u2} and \ref{u3}, except for $U1$, which has been slightly modified as follows:
\begin{equation} U1^*=\frac{\sum_{\forall i \in V_\Pi} v_i}{|V_\Pi|*vmax_p} \label{utility1*} \end{equation} 

The results for the generated problem set are shown in the plots from Figure \ref{fig:u3} to Figure \ref{fig:execution}. Each Figure consists of nine plots, resulting from the combination of the values '\textbf{F}ew', '\textbf{I}ndif' or '\textbf{M}any' for the preference $P_{\#visits}$  along the $X$ axis; and '\textbf{L}ow', '\textbf{I}ndif' or '\textbf{H}igh' for $P_{occupation}$ along the $Y$ axis. Each single plot pictures 4 bars, where the first corresponds to OPT-M1' and the other three correspond to CHO-M1 to CHO-M3. For example, the plot in the left-bottom corner of Figure \ref{fig:occup} shows, for each of the four plan metrics, the average percentage of occupation of the agendas for all the problems where the user has defined a '\textbf{H}igh' temporal occupation and a '\textbf{F}ew' number of visits.

\setlength{\tabcolsep}{5pt}

\begin{table}[tbp]

\centering
\caption{Average results wrt. the pref. of number of visits}
\label{tab:visits}
\begin{tabular}{|l|l|l|l|l|l|}
\hline
\multirow{2}{*}{\begin{tabular}[c]{@{}l@{}}Pref\\ \#Visit\end{tabular}} & \multirow{2}{*}{Metric} & \multicolumn{4}{l|}{Results}            \\ \cline{3-6}
                                                                        &                         & \#Visit & U1* & U2 & U3 \\ \hline
\multirow{4}{*}{Many}                                                   & OPT/M1'                 & 2,704   & 0,830   & 0,467    & 0,846    \\ \cline{2-6}
                                                                        & CHO/M1                  & 3,259   & 0,856   & 0,580    & 0,862    \\ \cline{2-6}
                                                                        & CHO/M2                  & 3,148   & 0,873   & 0,651    & 0,885    \\ \cline{2-6}
                                                                        & CHO/M3                  & 3,074   & 0,872   & 0,654    & 0,884    \\ \hline
\multirow{4}{*}{Indif.}                                                 & OPT/M1'                 & 2,444   & 0,836   & 0,468    & 0,850    \\ \cline{2-6}
                                                                        & CHO/M1                  & 3,185   & 0,857   & 0,574    & 0,865    \\ \cline{2-6}
                                                                        & CHO/M2                  & 2,463   & 0,910   & 0,682    & 0,918    \\ \cline{2-6}
                                                                        & CHO/M3                  & 2,556   & 0,893   & 0,670    & 0,902    \\ \hline
\multirow{4}{*}{Few}                                                    & OPT/M1'                 & 1,296   & 0,886   & 0,359    & 0,887    \\ \cline{2-6}
                                                                        & CHO/M1                  & 1,685   & 0,878   & 0,403    & 0,881    \\ \cline{2-6}
                                                                        & CHO/M2                  & 1,407   & 0,922   & 0,612    & 0,923    \\ \cline{2-6}
                                                                        & CHO/M3                  & 1,648   & 0,913   & 0,610    & 0,916    \\ \hline
\end{tabular}
\end{table}

Figures from \ref{fig:u3} to \ref{fig:u1} show the average utility measured by $U1^*$, $U2$ and $U3$ (equations \ref{utility1*}, \ref{u2} and \ref{u3}, respectively) for the 9 combinations of user preferences. We can observe that the utility values are very similar for the four metrics in Figures \ref{fig:u3} and \ref{fig:u1}. However, in Figure \ref{fig:u2}, the values are significantly lower for {\small{\sf OPTIC}}. This is because {\small{\sf CHOCO}}, which returns the optimal value for the three metrics, tends to principally minimize the penalty for the non-visited POIs of the list $V$. Consequently, {\small{\sf CHOCO}} plans will typically include the most-valued POIs, a larger number of POIs or the POIs that render more utility per unit time and so the plans will have higher utility values, which is specially notable for the '\textbf{L}ow' occupation. This is also confirmed with the results of Tables \ref{tab:occup} and \ref{tab:visits}. Table \ref{tab:occup} shows the average results of the utility measures in the solution plans obtained for the problem instances with occupation '\textbf{H}igh', '\textbf{I}ndif' and '\textbf{L}ow'. And Table \ref{tab:visits} shows the average results of the utility in the solution plans obtained for the problem instances where the user selected '\textbf{M}any', '\textbf{I}ndif' o '\textbf{F}ew' number of visits. We can also observe in the Tables that the average occupation and number of visits is always higher in {\small{\sf CHOCO}} for the aforementioned reason. For instance, for problems where the user selected '\textbf{F}ew' visits, {\small{\sf OPTIC}} includes only one POI in the majority of plans while {\small{\sf CHOCO}} includes two POIs.

The plot in Figure \ref{fig:occup} shows the percentage of occupation for each of the 9 possible combinations of user preferences. The four plan metrics yield high values of occupation when $P_{occupation}$ '\textbf{H}igh'. However, when the user selects '\textbf{L}ow' or '\textbf{I}ndif', we can observe the values of {\small{\sf OPTIC}} are significantly lower, which could be interpreted as a result more compliant with a '\textbf{L}ow' occupation (for the case '\textbf{I}ndif' any value would be equally acceptable). Likewise, this is explained because {\small{\sf CHOCO}} solution represents a plan in which the user is visiting highly-recommendable POIs for longer or the plan includes more POIs than the plans returned by {\small{\sf OPTIC}}, and this negatively affects the occupation in problems where the this preference is set to '\textbf{L}ow'. However, given that the values of {\small{\sf OPTIC}} plans are around 40\% of the route occupancy, one might see this as an 'extremely low' value. This interpretation would obviously depend on the user likes, an indication that a more accurate definition of user preferences might be preferable. On the other hand, CHO-M2 and CHO-M3 are clearly the metrics less compliant with a '\textbf{L}ow' occupancy preference, an indication that {\small{\sf CHOCO}} tends to maximize the time the user is visiting highly-recommendable POIs.

The plots in Figure \ref{fig:visits} show the average number of visits. In general, metrics OPT-M1' and CHO-M1 are the best performers, except in the '\textbf{F}ew-\textbf{H}igh' dimension and '\textbf{M}any-\textbf{L}ow' dimension, respectively. This reflects the fact that OPT-M1' is more sensible to the '{\bf L}ow' occupation preference, similarly to Figure \ref{fig:occup}, whereas CHO-M1 is more sensible to the '{\bf H}igh' occupation preference. With respect to the execution time shown in Figure \ref{fig:execution} (time in minutes), again OPT-M1' and CHO-M1 are the top performers in all the dimensions. In general, we can observe that when $P_{occupation}$ is '\textbf{H}igh' and $P_{\#visits}$ is '\textbf{M}any', both solvers require longer to solve the problems with any metric.

The values in Tables \ref{tab:occup} and \ref{tab:visits} allow us to examine the plan metrics with respect to the achievement of the user preferences. Table \ref{tab:occup} shows that the best metrics wrt the utility and level of occupation for a '\textbf{H}igh' and '{\bf I}ndif' $P_{occupation}$ are CHO-M2 and CHO-M3, although the differences in the '\textbf{H}igh' case are smaller. In the '{\bf L}ow' case, the best results are obtained by OPT-M1', followed by CHO-M1. This follows the same tendency as above, where {\small{\sf OPTIC}} is more compliant with '{\bf L}ow' occupation values. In Table \ref{tab:visits}, the best metric for '\textbf{M}any' and '{\bf I}ndif' number of visits is CHO-M1; however, CHO-M2 obtains better utility values. This means that, although CHO-M2 includes less visits in the plan, the utility per time unit for the user is higher. For the case of '\textbf{F}ew' visits, the best metrics is OPT-M1', in line with the rest of results, followed by CHO-M2, which obtains higher utility values. We can conclude that maximizing $\sum\limits_{p \in V_\Pi}(v_p*dur_p)$ provides the best utility to the user and that CHO-M2 is the best plan metric.

\section{Conclusion}

This paper describes the tourist problem, which consists in creating a personalized tourist agenda taking into account, apart from the usual constraints, such as maximizing the user satisfaction with the visits, other user preferences related to the travel style. We detail how this problem can be solved both using an automated planner and a CSP solver. We tested various plan metrics in two problem sets and showed that a plan metric that takes into account all the activities of the tour, including travelling times between places, yields in general a better utility to the user.

\section{Acknowledgments}
This work has been partly supported by the Spanish MINECO under project TIN2014-55637-C2-2-R.


\bibliographystyle{style/aaai}

\bibliography{b}

\begin{thebibliography}{}

\bibitem[\protect\citeauthoryear{Alexiadis and Refanidis}{2016}]{AlexiadisR16}
Alexiadis, A., and Refanidis, I.
\newblock 2016.
\newblock Alternative plan generation and online preference learning in
  scheduling individual activities.
\newblock {\em International Journal on Artificial Intelligence Tools}
  25(3):1--28.

\bibitem[\protect\citeauthoryear{Benton, Coles, and Coles}{2012}]{BentonCC12}
Benton, J.; Coles, A.~J.; and Coles, A.
\newblock 2012.
\newblock Temporal planning with preferences and time-dependent continuous
  costs.
\newblock In {\em Proc. ICAPS},  2--10.

\bibitem[\protect\citeauthoryear{Buhalis and Amaranggana}{2013}]{buhalis13}
Buhalis, D., and Amaranggana, A.
\newblock 2013.
\newblock Smart tourism destinations enhancing tourism experience through
  personalisation of services.
\newblock In {\em Proc. Int. Confernece on Information and Communication
  Technologies in Tourism},  553--564.

\bibitem[\protect\citeauthoryear{Castillo \bgroup et al\mbox.\egroup
  }{2008}]{castillo08}
Castillo, L.~A.; Armengol, E.; Onaindia, E.; Sebastia, L.;
  Gonz{\'{a}}lez{-}Boticario, J.; Rodr{\'{\i}}guez, A.; Fern{\'{a}}ndez, S.;
  Arias, J.~D.; and Borrajo, D.
\newblock 2008.
\newblock samap: An user-oriented adaptive system for planning tourist visits.
\newblock {\em Expert Systems with Applications} 34(2):1318--1332.

\bibitem[\protect\citeauthoryear{Chen, Wah, and Hsu}{2005}]{chen05}
Chen, Y.; Wah, B.~W.; and Hsu, C.
\newblock 2005.
\newblock Temporal planning using subgoal partitioning and resolution in
  {SGPlan}.
\newblock {\em Journal of Artificial Intelligence Research} 26:323--369.

\bibitem[\protect\citeauthoryear{Edelkamp and Hoffmann}{2004}]{edelkamp04}
Edelkamp, S., and Hoffmann, J.
\newblock 2004.
\newblock {PDDL2.2}: the language for the classical part of {IPC--4}.
\newblock In {\em Proc. International Planning Competition (ICAPS)},  2--6.

\bibitem[\protect\citeauthoryear{Edelkamp, Jabbar, and
  Nazih}{2006}]{Gerevini06-mips}
Edelkamp, S.; Jabbar, S.; and Nazih, M.
\newblock 2006.
\newblock Large-scale optimal {PDDL3} planning with mips-xxl.
\newblock In {\em 5th International Planning Competition Booklet},  28--30.

\bibitem[\protect\citeauthoryear{Fox and Long}{2003}]{fox2003pddl2}
Fox, M., and Long, D.
\newblock 2003.
\newblock {PDDL2.1}: An extension to {PDDL} for expressing temporal planning
  domains.
\newblock {\em Journal of Artificial Intelligence Research} 20:61--124.

\bibitem[\protect\citeauthoryear{Garcia \bgroup et al\mbox.\egroup
  }{2009}]{Sebastia:2009}
Garcia, I.; Sebastia, L.; Onaindia, E.; and Guzman, C.
\newblock 2009.
\newblock {e-Tourism: a tourist recommendation and planning application}.
\newblock {\em International Journal on Artificial Intelligence Tools}
  18(5):717--738.

\bibitem[\protect\citeauthoryear{Garrido, Arangu, and
  Onaindia}{2009}]{Garrido09}
Garrido, A.; Arangu, M.; and Onaindia, E.
\newblock 2009.
\newblock A constraint programming formulation for planning: from plan
  scheduling to plan generation.
\newblock {\em Journal of Scheduling} 12(3):227--256.

\bibitem[\protect\citeauthoryear{Gerevini \bgroup et al\mbox.\egroup
  }{2009}]{gerevini2009deterministic}
Gerevini, A.~E.; Haslum, P.; Long, D.; Saetti, A.; and Dimopoulos, Y.
\newblock 2009.
\newblock Deterministic planning in the fifth international planning
  competition: {PDDL3} and experimental evaluation of the planners.
\newblock {\em Artificial Intelligence} 173(5):619--668.

\bibitem[\protect\citeauthoryear{Hoffmann}{2003}]{Hoffmann03}
Hoffmann, J.
\newblock 2003.
\newblock The metric-ff planning system: Translating ''ignoring delete lists''
  to numeric state variables.
\newblock {\em J. Artif. Intell. Res. {(JAIR)}} 20:291--341.

\bibitem[\protect\citeauthoryear{Ib{\'{a}}{\~{n}}ez, Sebastia, and
  Onaindia}{2016}]{IbanezSO16}
Ib{\'{a}}{\~{n}}ez, J.; Sebastia, L.; and Onaindia, E.
\newblock 2016.
\newblock Planning tourist agendas for different travel styles.
\newblock In {\em Proc. ECAI},  1818--1823.

\bibitem[\protect\citeauthoryear{Kurata and Hara}{2014}]{kurata14}
Kurata, Y., and Hara, T.
\newblock 2014.
\newblock Ct-planner4: Toward a more user-friendly interactive day-tour
  planner.
\newblock In {\em Proc. Int. Information and Communication Technologies in
  Tourism},  73--86.

\bibitem[\protect\citeauthoryear{Lim \bgroup et al\mbox.\egroup
  }{2015}]{LimCLK15}
Lim, K.~H.; Chan, J.; Leckie, C.; and Karunasekera, S.
\newblock 2015.
\newblock Personalized tour recommendation based on user interests and points
  of interest visit durations.
\newblock In {\em Proc. 24th IJCAI},  1778--1784.

\bibitem[\protect\citeauthoryear{Prud'homme, Fages, and Lorca}{2016}]{choco}
Prud'homme, C.; Fages, J.-G.; and Lorca, X.
\newblock 2016.
\newblock {\em Choco Documentation}.
\newblock TASC, INRIA Rennes, LINA CNRS UMR 6241, COSLING S.A.S.

\bibitem[\protect\citeauthoryear{Refanidis and Alexiadis}{2011}]{RefanidisA11}
Refanidis, I., and Alexiadis, A.
\newblock 2011.
\newblock Deployment and evaluation of selfplanner, an automated individual
  task management system.
\newblock {\em Computational Intelligence} 27(1):41--59.

\bibitem[\protect\citeauthoryear{Refanidis \bgroup et al\mbox.\egroup
  }{2014}]{RefanidisE15}
Refanidis, I.; Emmanouilidis, C.; Sakellariou, I.; Alexiadis, A.; Koutsiamanis,
  R.-A.; Agnantis, K.; Tasidou, A.; Kokkoras, F.; and Efraimidis, P.~S.
\newblock 2014.
\newblock my{V}isit{P}lanner \({}^{\mbox{gr}}\): {P}ersonalized {I}tinerary
  {P}lanning {S}ystem for {T}ourism.
\newblock In {\em Proc. SETN},  615--629.

\bibitem[\protect\citeauthoryear{Rodr{\'{i}}guez \bgroup et al\mbox.\egroup
  }{2012}]{RodriguezMPC12}
Rodr{\'{i}}guez, B.; Molina, J.; P{\'{e}}rez, F.; and Caballero, R.
\newblock 2012.
\newblock Interactive design of personalised tourism routes.
\newblock {\em Tourism Management} 33(4):926--940.

\bibitem[\protect\citeauthoryear{Vansteenwegen \bgroup et al\mbox.\egroup
  }{2011}]{vansteenwegen11}
Vansteenwegen, P.; Souffriau, W.; Berghe, G.~V.; and Oudheusden, D.~V.
\newblock 2011.
\newblock The city trip planner: An expert system for tourists.
\newblock {\em Expert Syst. Appl.} 38(6):6540--6546.

\end{thebibliography}

\end{document}